# Correlated Non-Parametric Latent Feature Models

**Finale Doshi-Velez**
Cambridge University
Cambridge, UK

**Zoubin Ghahramani**
Cambridge University
Cambridge, UK

## Abstract

We are often interested in explaining data through a set of hidden factors or features. When the number of hidden features is unknown, the Indian Buffet Process (IBP) is a nonparametric latent feature model that does not bound the number of active features in dataset. However, the IBP assumes that all latent features are uncorrelated, making it inadequate for many realworld problems. We introduce a framework for correlated non-parametric feature models, generalising the IBP. We use this framework to generate several specific models and demonstrate applications on realworld datasets.

## 1 INTRODUCTION

Identifying structure is a common problem in machine learning. For example, given a set of images, we may wish to extract the objects that compose them. Slightly more abstractly, a country's development statistics might depend on its geographic location or political system. In the unsupervised setting, both the identities of the features—that is, if they correspond to objects or locations or some more complex aspect of the data—and assignments of features to observations are unknown. In many real-world situations, the number of hidden features is also unknown.

The Indian Buffet Process (Griffiths & Ghahramani, 2005) is an attractive prior on feature-assignments because it does not fix the number of hidden features realised in a set of observations. While the number of realised features in a finite dataset is guaranteed to be finite, the number of features is expected to grow with the number of observations. These properties are realistic in many situations: for example, we expect to see new objects as we observe more images.

The IBP prior assumes the observations exchangeable and the features are independent. Recent work has extended the IBP prior to include correlated observations. The phylogenetic IBP (Miller et al., 2008) uses a tree structure to describe similarities between observations, and similar observations are likely to contain similar features. For time series data, the infinite factorial HMM (Van Gael et al., 2009) is an IBP-like prior that encodes subsequent observations are likely to share features.

In many scenarios, the features are also correlated. For example, suppose observations correspond to image pixels. Latent features might be objects in the scene, such as a pen or lamp. Sets of objects—that is, certain latent features—may generally occur together: an image with a desk is likely to contain a pen; an image with a knife is likely to contain a fork. When modelling correlations, the correlated topic model (Blei & Lafferty, 2006) directly learns parameters for a joint distribution. An alternative approach might just group co-occurring objects as a single latent feature and thus avoid the need to model correlations between features. However, ignoring the underlying structure could result in less robust inference if the set of features does not always occur as a set—for example, if a pen is missing from a particular desk—and fails to leverage co-occurring features for inference.

Our approach, however, draws on the hierarchical structures of sigmoid (Neal, 1992) and deep belief (Hinton et al., 2006; Bengio, 2007) networks, where correlations reflect a higher layer of structure. For example, in the image scenario, being at a table may explain correlations between knives and forks. In the nonparametric setting, topic models such as Pachinko allocation (Li et al., 2007) also use hierarchies to model correlations. Closest to our interest in nonparametric feature models is infinite hierarchical factorial regression (Rai & Daume, 2009). Infinite hierarchical factorial regression (IHFR) is a partially-conditional model that uses an IBP to determine what features are present and then applies the coalescent to model correlations given the active features. Drawing on the use of hierarchies in topic models and deep belief nets, we develop a more general framework for unconditional nonparametric correlated feature models and demonstrate applications to several realworld datasets.

## 2 GENERAL FRAMEWORK

Our nonparametric correlated featured model places a prior over a structure describing the correlations and cooccurrences among an infinite number of features and observations. Inference on this infinite structure is tractable if the prior ensures finite set of observations affects only a finite part of the structure. More generally, the following properties would be desirable in a nonparametric correlated feature model:

- A finite dataset should contain a finite number of latent features with probability one.
- Features and data should remain exchangeable.
- Correlations should capture motifs, or commonly occuring sets of features.

The first desideratum requires particular attention if the hidden features are correlated. The prior must ensure that the correlations do not cause an infinite number of features to be expressed in any observation.

Let the feature assignment matrix $Z$ be a binary matrix where $z_{nk} = 1$ if feature $k$ is present in observation $n$. In our model, $Z$ depends on a set of category assignments $C$ and a set of category-feature relations $M$ (see graphical model in figure 1) . The binary category-assignment matrix $C$ contains $c_{nl} = 1$ if observation $n$ is a member of category $l$. Similarly, $m_{lk} = 1$ if feature $k$ is associated with category $l$. Features become correlated because observations choose features only through categories, and categories are associated with sets of features (see figure 2 for a more explicit illustration). Finally, the data $X$ are produced by the feature assignments $Z$ and some parameters $A$.

**Formal Description.** For each observation, the generative model first draws one or more categories via a non-parametric process with parameter $\alpha_C$.

$$C \sim \mathsf{NP1}(\alpha_C) \qquad (1)$$

where $c_{nl}$ indicates whether observation $n$ belongs to category $l$. A second nonparametric process with pa-

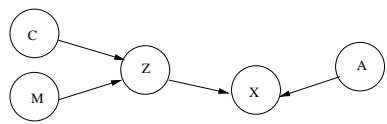

Figure 1: Graphical model. Feature assignments $Z$ depend on category assignments $C$ and category-feature relations $M$. Data $X$ depend on $Z$ and parameters $A$ (all hyperparameters omitted for clarity).

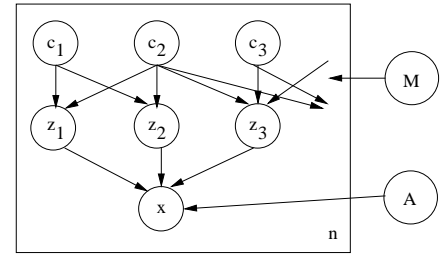

Figure 2: Plate for one observation. Observation $n$ selects a set of categories $c_{ni}$ which in turn select features $z_{nj}$. The connection matrix $M$ describes the links between features and categories.

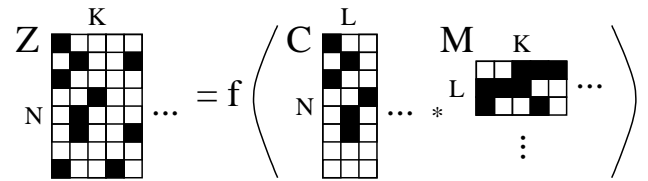

Figure 3: Cartoon of the matrix product. The function $f$ describes how feature assignments are derived from the category matrix $C$ and the connection matrix $M$.

rameter $\alpha_M$ associates categories with features:

$$M \sim \mathsf{NP2}(\alpha_M) \qquad (2)$$

where $m_{lk}$ indicates whether category $l$ chose feature $k$. The processes NP1 and NP2 should ensure that, with probability one, each observation is associated with a finite number of categories and each category is associated with a finite number of features. Finally, the feature-assignment matrix $Z = f(CM)$, where $f$ is some (possibly stochastic) function that converts the matrix product $CM$ into a set of binary values. Figure 3 shows a cartoon of this process.[1]

We summarise sufficient conditions on the binary-valued function $f$ to ensure a finite set of observations will contain a finite number of active features below:

**Proposition 1.** *If $z_{nk} = f(c_n^\top m_k)$ and $c_n^\top m_k = 0$ implies $f(c_n^\top m_k) = 0$, then the number of features in a finite dataset will be finite with probability one.*

*Proof.* Let $L$ be the number of active categories in $N$ observations, and let $K$ be the number of active features in the $L$ categories. The conditions on NP1 ensure $L$ is bounded with probability one. Since $L$ is bounded, the conditions on NP2 ensure $K$ is also bounded with probability one. Let $C_{1:N}$ denote the first $N$ rows of $C$. Thus the product $C_{1:N}M$ contains at most $NK$ nonzero values. The second condition on $f$ ensures that $Z_{1:N}$ has a finite number of non-zero values with probability one. □

---
[1] It is possible to add more layers to the hierarchy, but we believe two layers should suffice for most applications. Inference also becomes more complex with more layers.

Intuitively, the sufficient conditions imply (1) only the presence—not the absence–of a category can cause features to be present in the data and (2) categories can only cause the presence of the features associated with them. These implications are similar to those of the standard IBP model, where we require only the presence of a feature has an effect on the observations.

The previous discussion was limited to situations where $C$, $M$, and $Z$ are binary. However, other sparse processes may be used for $C$ and $M$, and the output of $f(\cdot)$ need not be binary as long as $f(0) = 0$ with probability one. For example, the infinite gamma-poisson process (Titsias, 2007) creates a sparse integer-valued matrix; such a prior may be appropriate if categories are associated with multiple copies of a feature.

## 3 SPECIFIC MODELS

Many choices exist for the nonparametric processes NP1 and NP2 and the function $f$. Here we describe nested models that use the Dirichlet Process (DP) and the Indian Buffet Process as base processes. However, other models such as the Pitman-Yor Process could also be used. The DP-IBP model is a factorial approach to clustering where we expect clusters to share features. The IBP-IBP models add an additional layer of sophistication: an observation may be associated with multiple feature sets, and sets may share features.

The DP is a distribution on discrete distributions which can be used for clustering (for an overview, see (Teh, 2007)). We represent the DP in matrix form by setting $c_{nl} = 1$ if observation $n$ belongs to cluster $l$. The IBP (Griffiths & Ghahramani, 2005) is a feature model in which each observation is associated with $\text{Poisson}(\alpha)$ features. Similar to the DP, a few popular features are present in most of the observations. Specifically, given $N$ total observations, the probability that observation $n$ contains an active feature $k$ is $r_k/N$, where $r_k$ is the number of observations currently using feature $k$. Both the DP and the IBP are exchangeable in the observations and features.

### 3.1 DP-IBP MODEL

The DP-IBP model draws $C$ from a Dirichlet Process and $M$ from an Indian Buffet Process. We let $f(c_n^\top m_k) = c_n^\top m_k$ and thus $Z = CM$:

$$\begin{aligned} C &\sim \text{DP}(\alpha_C) \\ M &\sim \text{IBP}(\alpha_M) \\ z_{nk} &= c_n^\top m_k \end{aligned} \quad (3)$$

In the context of the Chinese restaurant analogy for the DP, the DP-IBP corresponds to customers (observations) sitting at tables associated with combo meals (categories) instead of single dishes, and different combo meals may share specific dishes (features).

As in the DP, the dishes themselves are drawn from some continuous or discrete base distribution.

**Properties.** The properties of the DP and IBP ensure the DP-IBP will be exchangeable over features and the observations. The distribution over the number of features has no closed form, but we can bound its expectation. The expected number of categories $N_C$ in a DP with $N$ observations is $O(\alpha_C \log(N))$. Given $N_C$, the number of features $N_f$ is distributed as $\text{Poisson}(\alpha_M H_{N_C})$, where $H_{N_C}$ is the harmonic number corresponding to $N_C$. We apply Jensen's inequality to the iterated expectations expression for $E[N_f]$ to bound the expected number of features:

$$\begin{aligned} E[N_f] &= E_c[E_m[N_f|N_c]] \quad (4) \\ &= E_c[\alpha_M H_{N_c}] \\ &= E_c[\alpha_M \log(N_c) + O(1)] \\ &\leq \alpha_M \log(E_C[N_c]) + O(1) \\ &= \alpha_M \log(O(\alpha_C \log(N))) + O(1) \\ &= O(\log \log(N)) \end{aligned}$$

**Inference.** We apply the partial Gibbs sampling scheme described in (Neal, 2000) to resample the DP category matrix $C$. The IBP matrix $M$ can be resampled using the standard equations described in (Griffiths & Ghahramani, 2005). In both cases, the sampling equations have the same general form:

$$\begin{aligned} P(m_{lk}|X, Z, C, M_{-lk}, A) & \quad (5) \\ \propto P(m_{lk}|M_{-lk})P(X|Z, A)P(Z|C, M) \end{aligned}$$

where $M_{-lk}$ denotes the elements of $M$ excluding $m_{lk}$. An attractive feature of the DP-IBP model is that because $Z$ is a deterministic function of $C$ and $M$, the likelihood term $P(X|Z, A)P(Z|C, M)$ reduces to $P(X|C, M, A)$. Because the data is directly considered when sampling categories and connections, without $Z$ as an intermediary, the sampler tends to mix quickly.

**Demonstration.** We applied the Gibbs sampler to a synthetic dataset of 700 block images from (Griffiths & Ghahramani, 2005). The 6x6 pixel images contained four types of blocks, shown in the lower left quadrant of figure 4, which always cooccurred in specific combinations (lower right quadrant). We ran 3 chains for 1000 iterations with the DP-IBP model, using a likelihood model of the form $X = ZA + \epsilon$. The features $A$ had an exponential prior and $\epsilon$ was Gaussian white noise uncorrelated across observations and dimensions. All hyperparameters sampled using vague Gamma priors.

The top half of figure 4 shows a representative sample from the inference. The DP-IBP recovers that the images contain four types of blocks cooccurring in nine combinations. In particular, the DP-IBP hierarchy allows the inference to naturally discover the null-cluster, corresponding to no features being present, without additional parameters (as required for IHFR (Rai & Daume, 2009)). The sampler quickly converges near the true number of features and clusters (figure 5).

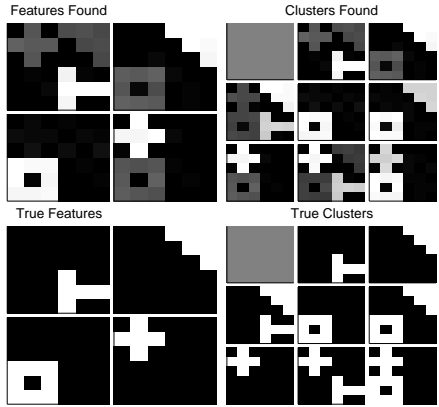

Figure 4: Sample showing structure found by the DP-IBP. Both the features and clusters (top row) closely match the underlying structure (bottom row).

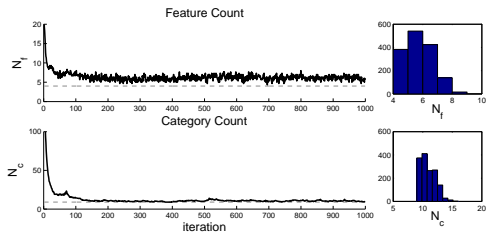

Figure 5: Evolution of the number of clusters and features (dashed lines show true values). The histograms of posterior over feature and cluster counts were computed from the final 500 samples. The hyperparameters converged in a similar fashion.

### 3.2 IBP-IBP MODEL

The DP-IBP associates each observation with only one cluster. However, some situations may naturally contain observations with memberships in multiple categories. For example, a image of a picnic may contain typical outdoor elements, such as trees and sky, as well as food-related objects. A multiple membership model at the category level would allow an observation to be part of multiple sets. In the IBP-IBP model, we place IBP priors on both $C$ and $M$ and set the link function $f$ to the 'or' of the product $c_n^\top m_k$:

$$
\begin{aligned}
C &\sim \mathsf{IBP}(\alpha_C) \\
M &\sim \mathsf{IBP}(\alpha_M) \\
z_{nk} &= (c_n^\top m_k) > 0
\end{aligned}
\qquad (6)
$$

**Properties.** The expected number of features in the IBP-IBP can be bounded similarly to the DP-IBP. The number of active categories $N_C$ in $N$ observations is distributed as $\mathsf{Poisson}(\alpha_C H_N)$, so the expected number of categories is still $O(\log(N))$. Given a number of categories, the number of features is distributed as $\mathsf{Poisson}(\alpha_M H_{N_C})$. Thus, by equation 4, the expected number of features is bounded by $O(\log \log(N))$. The distribution of $Z$ is exchangable in both the features and observations from the properties of the IBP.

**Inference.** To Gibbs sample in the IBP-IBP model, we use the equations of (Griffiths & Ghahramani, 2005) to sample both $C$ and $M$.

### 3.3 NOISY-OR IBP-IBP MODEL

The 'or' in the IBP-IBP model implies that a feature is present in an observation if any of its parent categories are present. This hard constraint may be unrealistic: for example, kitchen scenes may often contain refrigerators, but not always. The noisy-or IBP-IBP uses a stochastic link function in which

$$Pr[z_{nk} = 1] = 1 - q^{c_n^\top m_k} \qquad (7)$$

where $q \in [0, 1]$ is the noise parameter (Pearl, 1988).[2] An attractive feature of the noisy-or formulation is that the probability of a feature being present increases as more of its parent categories become active. For example, we might expect a scene tagged with both kitchen and dining categories may be more likely to contain a table than a scene tagged as only a kitchen.

The noisy-or IBP-IBP is summarised by:

$$
\begin{aligned}
C &\sim \mathsf{IBP}(\alpha_C) \\
M &\sim \mathsf{IBP}(\alpha_M) \\
z_{nk} &\sim \mathsf{Bernoulli}(1 - q^{c_n^\top m_k})
\end{aligned}
\qquad (8)
$$

**Properties.** The noisy-or IBP-IBP inherits its exchangeability properties and feature distribution from the IBP-IBP (the parameter $q$ only scales the expected number of features by a multiplicative constant). As $q \to 0$, the noisy-or IBP-IBP reduces to the IBP-IBP.

**Inference.** Because $f$ is stochastic, the feature assignments $Z$ must also be sampled. Given $M$, $C$, and $X$, the probability that a feature $z_{nk} = 1$ is given by

$$Pr[z_{nk} = 1] \propto (1 - q^{c_n^\top m_k}) Pr[x_n | z_n, A]. \qquad (9)$$

Gibbs sampling $C$ and $M$ is identical to the IBP-IBP case except the likelihood terms now depend on $Z$ and are generally independent of the data.[3] For example, when resampling $m_{lk}$,

$$P(m_{lk}|X, Z, C, M_{-lk}, A) \quad (10)$$
$$\propto P(m_{lk}|M_{-lk})P(X|Z, A)P(Z|C, M)$$
$$\propto P(m_{lk}|M_{-lk})P(Z|C, M)$$

---

[2] Proposition 1 requires that the noisy-or does not also leak, that is, $Pr[z_{nk} = 1]$ must be 0 if all the parents $c_n$ of $z_{nk}$ are zero.

The constraints of the noisy-or model pose problems when naively sampling a single element of $C$ or $M$. For example, suppose $n$ is the only observation using category $l$. According to the IBP prior, $Pr[c_{nl} = 1] = 0$. However, if $c_{nl}$ is the only active parent of feature $z_{nk}$, and $z_{nk} = 1$, then according to the likelihood $Pr[c_{nl} = 1] = 1$. In such situations, $c_{nl}$ and its children from $z_n$ should be sampled jointly.

Another problem arises when all the parents of $z_{nk}$ are inactive but the likelihood $P(x_n|z_n, A)$ prefers $z_{nk} = 1$. To set $z_{nk} = 1$, one of $z_{nk}$'s parents must become active. However, if $z_{nk} = 0$, $z_{nk}$'s parents are unlikely to turn on. While slow, jointly sampling $z_{nk}$ with its parents resolves this issue.

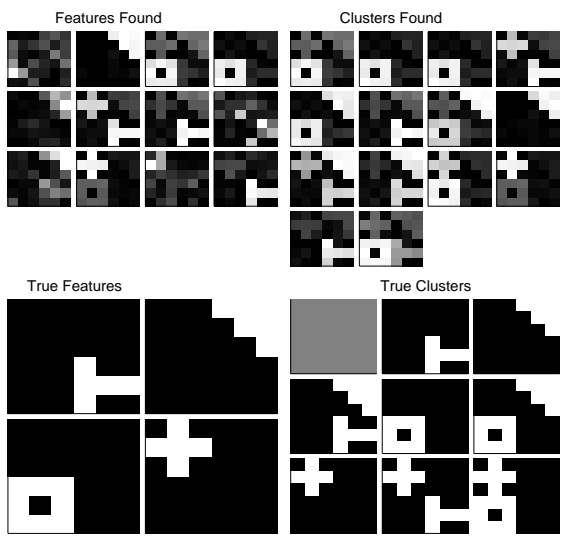

Figure 6: Sample showing structure found by the Noisy-Or IBP-IBP. Both the features and clusters (top row) reflect the underlying structure (bottom row), but often contain replicas.

**Demonstration.** We return to the blocks example of section 3.1, using the same clusters as before. However, unlike in section 3.1, we allow multiple (often overlapping) clusters to be present in an observation when generating the observations. Thus, the observations used to demonstrate the noisy-or IBP are *not* the same as the observations used in section 3.1; they have significantly more complex structure.

---
[3]The exception is when new features are being sampled.

Figure 6 shows the inferred features and clusters for a typical sample from the inference. The inferred features largely match the true features, but they are more noisy, and features are sometimes repeated. Similarly, the inferred clusters contain the true clusters and some replicas. The ghosted features and replicas are a common occurrence when sampling in IBP-like models; they occur when multiple observations propose similar features. Over time they tend to disappear, but this time can be exponential in the size of the dataset.[4]

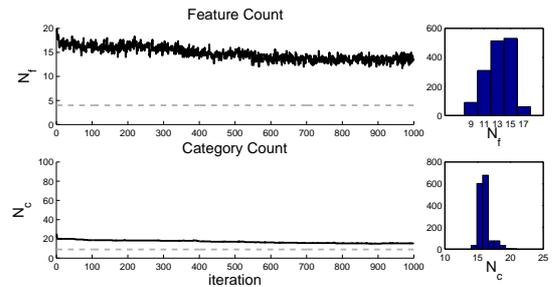

Figure 7: Evolution of the number of clusters and features. The dashed lines show the true values; the histograms show the posterior over the number of features and clusters. As before, the hyperparameters, given vague priors, converged in a similar fashion.

Figure 7 confirms that the number of features and categories is often overestimated. The posterior of the flexible noisy-or has many local optima; even after the sampler has mixed, many modes must be explored. However, the replicas do not prevent the noisy-or IBP-IBP from producing good reconstructions of the data.

## 4 EXPERIMENTS

We applied the three models of section 3 to five real-world datasets (table 1). The gene data consisted of expression levels for 226 genes from 251 subjects (Carvalho et al., 2008). The UN data consisted of a dense subset of global development indicators, such as GDP and literacy rates, from the UN Human Development Statistics database (UN, 2008). Similarly, the India dataset consisted of development statistics from Indian households (Desai et al., 2005). The joke data consisted of a dense subset of continuous-valued ratings of various jokes (Goldberg et al., 2001). Finally, the robot data consisted of hand-annotated image tags of whether certain objects occurred in images from a robot-mounted camera (Kollar & Roy, 2009).

Inference was performed using uncollapsed Gibbs sampling on 3 chains for 1000 iterations. Chains for more complex models were initialised with the final sample outputted by simpler models: the IBP and the DP-IBP were initialised with the output of the DP, the IBP-IBP was initialised from the DP-IBP, and the noisy-or IBP-IBP was initialised from the IBP-IBP. The likelihood model $x_n = z_n A + \epsilon$, where $a_{kd} \sim \text{Exponential}(\lambda)$ and the noise $\epsilon \sim \text{Normal}(0, \sigma_n)$ was used for the continuous-valued datasets. Under this model, the conditional posterior on $a_{kd}$ was a truncated Gaussian. For the binary robot data, the likelihood was given by

$$P(x_{nd} = 1 | z_{nd} = 1) = 1 - m_d$$
$$P(x_{nd} = 1 | z_{nd} = 0) = f_d$$

where $f_d$ was the probability of a false detection, and $m_d$ was the probability of a missed detection. These simple likelihood models are not necessarily the best match for complex, realworld data. However, even these simple models allowed us to find shared structure in the observations. (For a real application, of course, one would use an appropriately designed likelihood model.) Finally, all hyperparameters were given vague priors and sampled during the inference.

---
[4]One might introduce Metropolis moves to merge features or clusters. However, we find that these moves, while effective on small problems, have little effect in more complex, higher-dimensional realworld models.

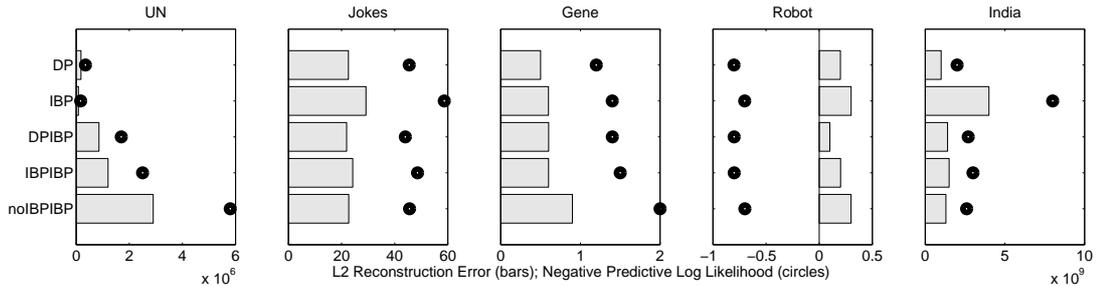

Figure 8: Negative predictive log-likelihoods (circles) and L2 reconstruction errors (bars) for held-out data from the realworld datasets. Both metrics have the same scale. Smaller values indicate better performance; note that models have consistent performance with respect to both predictive log-likelihoods and L2 reconstruction errors across all datasets.

**Quantitative Evaluation** We evaluated inference quality by holding out approximately $10D$ $X_{nd}$ values during the initial inference. No observation had all of its dimensions missing, and no dimension had all of its observations missing. Because models had different priors, inference quality was measured by evaluating the test log-likelihood and the L2 reconstruction error of the missing elements (a complete Bayesian model comparison would require an additional prior over the model classes). The evaluation metrics were averaged from the final 50 samples of the 3 chains.

Table 2 and figure 4 compare the reconstruction errors and predictive likelihoods of the three variants with the standard IBP, which does not model correlations between features, the DP, a flat clustering model. As we are most interested in feature-based models (IBP, DP-IBP, IBP-IBP, or noisy-or IBP-IBP), both the best-performing feature-based model, as well as the best overall model, are highlighted in bold. Plots for the hyperparameters are not shown, but the posteriors did converge during the inference.

Table 2: Predictive likelihoods on held-out data (higher is better). Bold figures indicate the best feature-based and best overall models for each dataset. Performance with respect to L2 reconstruction errors had similar patterns (see figure 4).

| Model | UN | Jokes | Gene | Robot | India |
|---|---|---|---|---|---|
| DP | -3.5e5 | -45.5 | **-1.2** | 0.8 | **-2.0e9** |
| IBP | **-1.7e5** | -58.7 | -1.5 | 0.7 | -8.0e9 |
| DPIBP | -17.0e5 | **-44.0** | -1.4 | **0.8** | -2.7e9 |
| IBPIBP | -25.0e5 | -48.6 | -1.6 | 0.8 | -3.0e9 |
| noIBPIBP | -58.0e5 | -45.6 | -2.0 | 0.7 | **-2.6e9** |

The structured variants outperformed the standard IBP in almost all cases. In particular, the DP-IBP usually had the best performance among the feature-based models. Its performance was on par with the DP—a simpler model with a much more robust inference procedure. (Indeed, we believe that the robustness of inference in the DP—and the difficulty of inference in the complex posterior landscapes of the structured models, is one of the main reasons for the difference in the models' performance.) Unlike the flat clusters provided by the DP, the DP-IBP can provide a structured representation of the data alongside good quantitative performance. These more qualitative benefits are ex-plored in the next section.

Table 1: Descriptions of data sets.

| Dataset | N | D | Description |
|---|---|---|---|
| UN | 155 | 15 | Human development statistics for 155 countries |
| Joke | 500 | 30 | User ratings (continuous) of 30 jokes |
| Gene | 251 | 226 | Expression levels for 226 genes |
| Robot | 750 | 23 | Visual object detections made by a mobile robot |
| India | 398 | 14 | Socioeconomic statistics for 398 Indian households |

**Qualitative Examples**[5] In table 2, we saw DP clustering had lower error rates than any of the feature-based models on the UN dataset. However, even in this case, the structured representation of the correlated feature models can provide some explanatory power.[6] The image in figure 9 shows a representative feature matrix $A$ from the DP-IBP. Each row in the matrix corresponds to a development statistic; for visualisation the values in each row have been scaled to $[0,1]$.

Columns correspond to features. We see that the first column, with high GDP and healthcare scores, is what we would expect to find in highly developed countries. The third column is representative of countries with an intermediate level of development, and the final column, with low GDP and high tuberculosis rates, has a pattern common in developing countries. The other columns highlight specific sets of statistics: column two augments technology and education, while column four corresponds to higher private healthcare spending and higher prison populations (and always occurs in conjunction with some other features).

What distinguishes the DP-IBP from simple clustering is that the features are shared among clusters. The circles on top of the feature matrix in figure 9 represent categories or clusters (only 5 of the 15 are shown). The $M$ matrix is visualised in the links between the categories and the features. Each column represents a feature, where the rows are the dimensions of the data. As with flat clustering, some categories ($C2, C5$) only use one feature (that is, connect to only one column). For example, certain developed nations such as Canada and Sweden are well characterised by only the first feature, and developing nations such as Mali and Niger are characterised by only the final feature.

However, feature sharing allows other categories to use multiple features. For example, the United States has high development statistics, like Canada and Sweden, but it also has, added on, relatively high tuberculosis rates, private healthcare spending, and prison populations. Showing what characteristics the United States shares with other countries is more informative than simply placing it in its own cluster.

The DP-IBP also found informative clusters in the robot data. As with the UN dataset, the DP-IBP was able to do a more sophisticated clustering that reflected shared elements in the data. Some of the categories of image tags discovered are listed below:

---
[5]We focus on the UN and robot datasets here; examples from other data sets will be made available online.
[6]We stress that as we are using unsupervised methods, the structures found cannot be considered to be discovering the 'true' or 'real' structure of dataset. We can only say that the structures found explain the data.

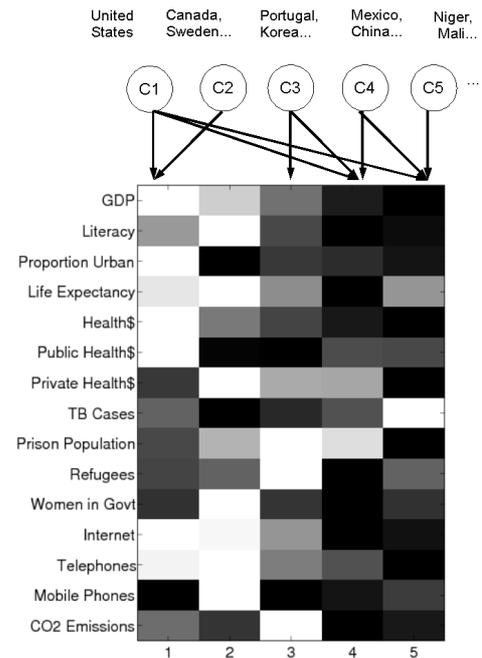

Figure 9: Part of a DP-IBP sample from the UN data. The rows in the image correspond to the development statistics, and columns represent feature vectors. The top row of circles, with links to various features, are some of the categories in this sample (there were 15 categories in total). Representative countries are listed above each category.

```
C1: hallway, door, trash can, chair, desk,
    office, computer, whiteboard
C2: door, trash can, robot, bike,
    printer, couch
C3: trash can, monitor, keyboard, book,
    robot, pen, plant
C4: hallway, door, trash can
```

Categories C1 and C4 often occurred singly in the data, in the form of simple clusters that corresponded to hallway and office scenes. Categories C2 and C3 often augmented category C1, reflecting office scenes that also included parts of a printing area (C2) and close-up views of desk spaces (C3).

## 5 DISCUSSION

We find in particular that the DP-IBP model, combining the unbounded number of clusters of a DP mixture with the nonparametric shared features of an IBP, provides a promising method for hierarchical hidden representations of data. Occupying a regime between pure clustering and pure feature-based modelling, the DP-IBP can capture dominant categorical qualities of the realworld datasets but still discover shared structure between clusters. It outperforms the standard IBP because allowing categories to share features lets each feature raw evidence from more observations and thus grow more refined. At the same time, forcing observations to be associated with only one category limits the model's flexibility. Thus, the DP-IBP has fewer of the identifiability issues common to feature-based models and produces more relevant categories.

In some situations, the more complex structured models may have been better matches for the data—for example, the robot data almost surely contained situations where multiple categories of noisy features were present. Here, better inference techniques could have proved beneficial. Split-merge moves may help accelerate mixing, but from a limited set of experiments we found that the benefits were highly data dependent: such moves provided some benefit in the toy images data set, where replica features tended to be a problem, but proved to be less useful in the more complex local optima of the real world data. Inference that uses "soft" assignments—such as variational techniques—may prove to be more robust to these local optima.

We see an important trade-off when choosing what kind of nonparametric model to apply. Our work was initially motivated to create a model for scenarios like the robot data, and thus we desired a generative process that would explain noisy, multiple-membership correlated feature models. An interesting question is how one may perform model selection across different choices of nonparametric priors within this general framework: while these models perform well when the prior reflects the data—such as in the toy blocks examples—the structure appropriate for realworld data is much more difficult to ascertain.

## 6 CONCLUSION

Our principled setting allows for an unbounded number of features and categories and provides a general framework for modelling a variety of different types of correlations. The framework can also model other useful properties such as feature sparsity (that is, many observations being feature-less). Interesting extensions might include incorporating aspects of the phylogenetic IBP or infinite factorial HMM to create models that consider correlations both between features and also between observations, as well as exploring methods to model negative correlations. The work in this paper also provides avenues which could be used to develop "deep" nonparametric models with multiple unbounded layers of hidden variables.

In particular, the DP-IBP model, combining nonparametric clustering with the shared features of an IBP, is a promising method for layered representations of latent variables. However, given the complexities of performing inference in these models, more analysis is needed to study the behaviours these models in real-world applications and to determine the sensitivities of the models to various hyper-priors. The work in this paper is only one step toward achieving more structured nonparametric models.

**Acknowledgements**

FD was supported by a Marshall scholarship.